\documentclass[smallcondensed]{svjour3}

\usepackage[T1]{fontenc}
\usepackage{amsfonts}
\usepackage{lmodern}
\usepackage{longtable}
\usepackage{float} %Cria opção H maiúsculo para não dar float em tabelas e figuras
\usepackage{setspace}                   % espaçamento flexível
\usepackage{makeidx} 
\usepackage{amsmath,amssymb}
\usepackage{relsize}
\usepackage[nottoc]{tocbibind}          % acrescentamos a bibliografia/indice/conteudo no Table of Contents
\usepackage{courier}                    % usa o Adobe Courier no lugar de Computer Modern Typewriter
\usepackage{type1cm}                    % fontes realmente escaláveis
\usepackage{listings}                   % para formatar código-fonte (ex. em Java)
\usepackage{titletoc}
\usepackage{multirow}
\usepackage{listings}
\usepackage{hyperref}
\hypersetup{
    colorlinks,
    citecolor=black,
    filecolor=black,
    linkcolor=black,
    urlcolor=black
}

\usepackage{algorithm}% http://ctan.org/pkg/algorithm
\usepackage{algpseudocode}% http://ctan.org/pkg/algorithmicx
\renewcommand{\algorithmicrequire}{\textbf{\small Input:}}
\renewcommand{\algorithmicensure}{\textbf{\small Output:}}

\usepackage{xpatch}
\xpatchcmd{\algorithmic}{\setcounter}{\algorithmicfont\setcounter}{}{}
\providecommand{\algorithmicfont}{}

\usepackage[font=small,format=plain,labelfont=bf,up,textfont=it,justification=justified,singlelinecheck=false]{caption}
\usepackage[usenames,svgnames,dvipsnames]{xcolor}

\usepackage{enumitem} %aplica estilos sobre os enumerate
\usepackage{graphicx} %imagens  não conversa com o \usepackage[pdftex]
\usepackage{amsmath} %subequation
\usepackage[a4paper,margin=1in,footskip=0.25in]{geometry}
\usepackage{multicol}
\usepackage{placeins} %\FloatBarrier uma alternativa ao \clearpage
\usepackage{savefnmark}
\usepackage{booktabs}
\usepackage{multicol}
\newcommand{\izbicki}[1]{}

\renewcommand{\vec}[1]{\mathbf{#1}}
\def\x{{\vec{x}}}

\def\X{{\vec{X}}}

\def\z{{\vec{z}}}
\def\Y{{\vec{Y}}}

\def\I{{\mathbb{I}}}
\def\E{{\mathbb{E}}}
\def\Sc{{S^c}}
\def\sF{{\mathcal{F}}}

\def\sY{{\mathcal{Y}}}

\def\sA{{\mathcal{A}}}
\def\bX{{\mathbb{X}}}
\def\bZ{{\mathbb{Z}}}
\def\ourtest{{COINP} }

\usepackage{array}
\newcolumntype{H}{>{\setbox0=\hbox\bgroup}c<{\egroup}@{}}

\usepackage{listings}
\newsavebox{\LstBox}

\usepackage{tikz}
\usetikzlibrary{positioning,shapes,arrows}

\usepackage[style=chicago-authordate,natbib=true,doi=false,url=false,hyperref=true]{biblatex}
\newtheorem{Def}{Definition}

\begin{document}

\title{Conditional independence testing: a predictive perspective}

\author{Marco Henrique de Almeida In\'acio \and
Rafael Izbicki \and
Rafael Bassi Stern
}

\institute{M. In\'acio \at
              Department of Statistics - Federal University of São Carlos (UFSCar) and \\ Institute of Mathematics and Computer Sciences - University of São Paulo (USP)\\
              \email{m@marcoinacio.com}           %  \\
           \and
           R. Izbicki \at
             Department of Statistics - Federal University of São Carlos (UFSCar)
             \and
          R. Stern \at
             Department of Statistics - Federal University of São Carlos (UFSCar)
}

%\date{Received: date / Accepted: date}
\date{}
% The correct dates will be entered by the editor

\maketitle
\vspace{-2cm}

\begin{abstract}
Conditional independence testing is a key problem required by
many machine learning and statistics tools. In particular, it is 
one way of evaluating the usefulness
of some features on a supervised prediction problem. We propose a novel conditional independence test in a predictive setting, and show that it achieves better power than competing approaches in several settings. Our approach consists in deriving a p-value using a permutation test where the predictive power using the unpermuted dataset is compared with the predictive power of using dataset where the feature(s) of interest are permuted. We conclude that the method achives sensible results on simulated and real datasets.\izbicki{...}
\end{abstract}

\section{Introduction}

Conditional independence testing is a key problem required by
many machine learning and statistics tools,
including Bayesian networks
\citep{jensen1996,campos2006}, time series \citep{DIKS20061647} \izbicki{referencias?}, causal inference \citep{spirtes2000,pearl2009} and feature selection \citep{koller1996}. 

Unfortunately, it is not possible to design
 conditional independence tests that  are powerful against all points in the alternative hypothesis \citep{shah2018}. Nevertheless, 
this issue can be partially addressed by assuming additional structure on the data distribution. Indeed,
various conditional independence methods
take advantage of specific settings to obtain improved power for alternatives of interest; see, for instance,
\citet{doran2014,sen2017,berrett2018,chalupka2018fast}
and references therein.
 
In this work we are interested
in testing conditional independence as a way of evaluating the usefulness
of some features on a supervised prediction problem.
More precisely, let $\X^S \subset (X_1,\ldots,X_p)$ be a subset of the features.
Our goal is to test if
$\X^S$ is independent of the label $Y$
conditionally on $\X^\Sc$, the remaining variables, i.e, we wish to test the hypothesis
$H_0: \X^S \bot Y | \X^\Sc$.

This is setting is closely connected to the literature of designing effective \emph{variable importance measures}, in which the goal
is to design indices that can be used to rank features according to how useful they are for predicting $Y$. A popular measure of variable
importance was designed by 
\citet{breiman2001}, but several alternative procedures are also available; see, for instance, \citet{strobl2008,fisher2018} and references therein.
Conditional independence testing is distinct from designing  importance measures
in the sense that its
 goal is not to quantify how
informative a given feature is, but instead to answer the question: ``is this feature relevant?".

In this work we propose
an approach to answer this question that consists in comparing the  performance of two prediction methods: the first is trained using all features, while the second  uses noise instead of the variables $S$.
We show that our method yields  a formal statistical hypothesis that approximately controls the  significance level, and that it achieves considerable power against
relevant alternative hypotheses.

Our work is related to \citet{watson2019}, who 
also compare the risk of
two prediction methods in order to test $H_0$. The main difference between these methods
is that
we use a permutation test-based statistic to compute p-values (see Section \ref{sec:experiments} for further details).
We show that this leads to substantial gain of power in several settings, and also a better control of type I error probabilities, especially for smaller
sample sizes.

The remaining of this paper is organized as follows.
Section \ref{sec:ourmethod}
introduces COINP, our approach
to test conditional independence.
Section \ref{sec:experiments}
contains experiments for comparing COINP with 
other approaches while Section \ref{sec:real_data} presents an illustrative example of applying the method to a real world dataset together the classical importance measure obtained from random forests.
Section \ref{sec:final} concludes the paper with final remarks.

\section{Conditional Independence Predictive Test (COINP)}
\label{sec:ourmethod}

\subsection{Notation and problem setting} 

Let $\mathcal{X}$ denote the feature space and 
$\mathcal{Y}$ the label space.
The observed data is $\bZ=(\bX,\Y)$,
where $\bX \in \mathcal{X}^n$ is
a  $n \times p$ feature matrix and
$\Y \in \mathcal{Y}^n$ is the label vetor.
We assume that the observations
$\z_i=(\x_i,y_i)$, $i=1,\ldots,n$,
are independent and identically distributed.  
Let $\X^S \subset (X_1,\ldots,X_p)$ be a subset of the features.
Our goal is to test if
$\X^S$ is independent of $Y$
conditionally on $\X^\Sc$, the remaining variables, i.e, we wish to test the hypothesis
$H_0: \X^S \bot Y | \X^\Sc$.

We denote by $\mathcal{F}:=\{f:\mathcal{X}\longrightarrow \mathcal{Y}\}$ the space of all mappings from features to outcomes (i.e., all prediction functions), and by
$\mathcal{Z}$ the set of all datasets. A prediction method
(e.g., neural networks or random forests)
is a function in the space
$\mathcal{A}=\{a:\mathcal{Z} \longrightarrow \mathcal{F}\}$.
We denote a loss function by
$L: \sY \times \sY \longrightarrow \mathbb{R}$. 
The risk  of a prediction function $f \in \sF$
is $R(f):=\E[L(f(\X),Y)]$.
We denote by 
$$\widehat{R}(f,\tilde{\bZ}):=\frac{1}{n}\sum_{i=1}^n L(f(\tilde{\x}_i),\tilde{y}_i)$$
the estimate of the risk of $f\in \sF$ based on a holdout dataset $\tilde{\bZ}$
of size $n$ (i.e., a dataset not used for obtaining $f$).

\begin{figure}[!h]
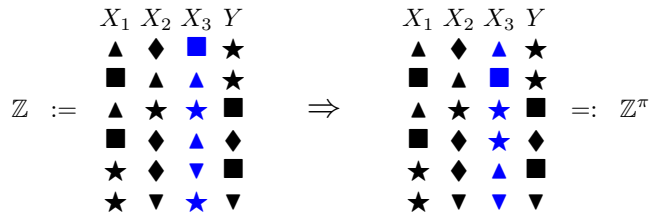

 \centering
 $$
 \bZ\hspace{2mm}:=\hspace{2mm}
\begin{matrix} 
X_1 & X_2 & X_3 & Y \\
\blacktriangle  & \blacklozenge & {\color{blue}\blacksquare} & \bigstar \\ 
\blacksquare  & \blacktriangle & {\color{blue}\blacktriangle} & \bigstar \\
\blacktriangle  & \bigstar & {\color{blue}\bigstar }& \blacksquare \\
\blacksquare  & \blacklozenge & {\color{blue}\blacktriangle} & \blacklozenge \\
\bigstar  & \blacklozenge & {\color{blue}\blacktriangledown} & \blacksquare \\
\bigstar  & \blacktriangledown & {\color{blue}\bigstar} & \blacktriangledown \\
\end{matrix} \hspace{8mm} \mathlarger{\mathlarger{ \Rightarrow}} 
\hspace{8mm}
\begin{matrix} 
X_1 & X_2 & X_3 & Y \\
\blacktriangle  & \blacklozenge & 
{\color{blue}\blacktriangle} & \bigstar \\ 
\blacksquare  & \blacktriangle & {\color{blue}\blacksquare} & \bigstar \\
\blacktriangle  & \bigstar & 
{\color{blue}\bigstar} & \blacksquare \\
\blacksquare  & \blacklozenge & {\color{blue}\bigstar }& \blacklozenge \\
\bigstar  & \blacklozenge & 
{\color{blue}\blacktriangle}& \blacksquare \\
\bigstar  & \blacktriangledown & {\color{blue}\blacktriangledown }
& \blacktriangledown \\
\end{matrix}\hspace{2mm}=:\hspace{2mm}\bZ^\pi
$$
\caption{Permutation procedure used to test if the $X_3$ is independent of $Y$ conditionally on 
 $X_1$ and $X_2$.}
\label{fig:perm}
\end{figure}

The Conditional Independence Predictive Test (COINP) requires one
to randomly permuting the rows of $\bX$ associated to the features $S$.
This procedure is illustrated in Figure \ref{fig:perm} for an example with $S=\{X_3\}$.
For every
$j=1,\ldots,B$, denote by $\bZ^{\pi_j}$ 
the $j$-th dataset obtained by performing this procedure on $\bZ$, and let $\x^{\pi_j}_i$ be the $i$-th observation of such dataset.
Similarly, we denote by 
$\tilde{\bZ}^{\pi_j}$  the $j$-th permutation of the  holdout set
$\tilde{\bZ}$.
Table \ref{tab:notation} summarizes the notation used in this paper.

\begin{table}[!h]
 \centering
 \caption{Notation used in the paper.}
 \begin{tabular}{l|l}
  \hline 
  Symbol
  & Meaning  \\ \hline 
  $\bX$ & $n \times p$ feature matrix \\
   $Y$ & Label vector \\
  $\bZ$ &Training data \\ 
  $\tilde{\bZ}$& Holdout data \\
  $\sF$ & space of all prediction functions \\
  $\sA$& space of all prediction methods \\
  $\widehat{R}(f,\bZ)$ & Estimate of the risk of $f \in \sF$ based on $\bZ$ \\
 $\bZ^{\pi_j}$ & $j$-th dataset with the rows of the $S$ columns of $\bZ$ are randomly permuted  \\
 $\x_i^{\pi_j}$ & The features of the  $i$-th row of $\bZ^{\pi_j}$ \\
  \hline 
 \end{tabular}
 \label{tab:notation}
\end{table}

\subsection{Proposed method}

The \ourtest procedure consists in  testing
$H_0$ by computing the rank of 
    $\widehat{R}(a(\bZ),\tilde{\bZ})$ among
    $$\{\widehat{R}(a(\bZ^{\pi_1}),\tilde{\bZ}^{\pi_1}),\ldots,\widehat{R}(a(\bZ^{\pi_B}),\tilde{\bZ}^{\pi_B})\},$$
where $a \in \sA$ is any prediction method.
We then reject the 
null hypothesis if this statistic is small.
The intuition
of this procedure
is that, if $H_0$ does not hold (i.e., if the features $S$ still have information about $Y$ even given $S^c$), an algorithm that uses $S$ in addition to $S^c$ will result in greater predictive power than one that does not use those features.
Thus,
permuting the rows associated with $S$ will result in a prediction function with larger risk. Hence, if $H_0$ does not hold, 
$\widehat{R}(a(\bZ),\tilde{\bZ})$ should be smaller than most 
$\widehat{R}(a(\bZ^{\pi_j}),\tilde{\bZ}^{\pi_j})$'s. 
On the other hand,
if
$H_0$ holds, the features $\X^S$ bring no additional gain in the predictive performance of 
$a$. It follows that the estimated risk
$\widehat{R}(a(\bZ),\tilde{\bZ})$
should be identically distributed
to $\widehat{R}(a(\bZ^{\pi_j}),\tilde{\bZ}^{\pi_j})$.
It follows that the rank statistic is uniformly distributed under the null hypothesis. \izbicki{precisa deixar mais preciso; acho que que identicamente distribuido eh sempre verdade (mas tem uns porens), mas nao necessariamente ha independencia}
This justifies the following COINP procedure, which we formally state in the following definition.

\begin{Def}[\ourtest-- Conditional Independence Predictive Test] Let $\alpha \in (0,1)$ and
$a \in \sA$ be a predictive method. The $\alpha$-level
Conditional Independence Predictive Test consists in
rejecting the null hypothesis $H_0: \X^S \bot Y | \X^\Sc$ if, and only if,
\begin{align}
\label{eq:perm_pval}
\frac{1}{B}\sum_{j=1}^B \I\left(\widehat{R}(a(\bZ),\tilde{\bZ}) \geq \widehat{R}(a(\bZ^{\pi_j}),\tilde{\bZ}^{\pi_j})\right) \leq \alpha.
\end{align}

\end{Def}

The \ourtest is described in 
Algorithm \ref{alg:perm}. Notice that
the  left-hand side 
of Equation \ref{eq:perm_pval} is in fact
a p-value based on  a permutation test \citep{good2013} \izbicki{develop?}.

\begin{algorithm}
  \caption{ \small \ourtest}\label{alg:perm}
  \algorithmicrequire \ {\small  
  training data $\bZ$,
  testing data $\tilde{\bZ}$,
  prediction method
  $a \in \sA$, loss function
  $L$, feature indices $S$, number of simulations $B$ } \\
  \algorithmicensure \ {\small p-value  for testing 
   $H_0:  \X^S \bot Y | \X^\Sc$}
   
\begin{algorithmic}[1]
    \State $f\gets a(\bZ)$
    \State 
  $R\gets\widehat{R}(f,\tilde{\bZ})$ 
    \For{$j \in \{1,\ldots,B\}$}
        \State Compute $\bZ^{\pi_j}$ by randomly permuting the columns of $\bZ$ associated to features $S$ 
        \State $f_j \gets a(\bZ^{\pi_j})$
      \State Compute $\tilde{\bZ}^{\pi_j}$ by randomly permuting the columns of $\tilde{\bZ}$ associated to features $S$ 
      \State $R_j\gets\widehat{R}(f_j,\tilde{\bZ}^{\pi_j})$
    \EndFor
    \State \textbf{return}  
    $|\{j: R \geq R_j\}| / B$
  \end{algorithmic}
\end{algorithm}

\section{Experiments}
\label{sec:experiments}

\subsection{Other approaches}

We compare our permutation method with the following approaches.

\subsubsection{Conditional Predictive Impact (CPI)}
  
  The
  Conditional Predictive Impact (CPI) test, introduced by \citet{watson2019},
  consists in training two prediction methods: one on the original dataset, $f=a(\bZ)$, and another one on a permuted dataset,
  $f_1=a(\bZ^{\pi_1})$. It then tests the one-sided null hypothesis 
    $H_0:R(f,\tilde{\bZ})\geq R(f_1,\tilde{\bZ}^{\pi_1})$
    by checking if the distribution of the loss function on the original set,
    $$(L(f(\tilde{\x}_1),\tilde{y}_1),\ldots,L(f(\tilde{\x}_n),\tilde{y}_n))$$ 
    comes from a distribution with
    smaller average than the 
  distribution of the loss function on the permuted test,
    $$(L(f_1(\tilde{\x}^{\pi_1}_1),\tilde{y}_1),\ldots,L(f_1(\tilde{\x}^{\pi_1}_n),\tilde{y}_n)).$$
  In practice, we use the paired
    t-test to perform this comparison; see \citet{watson2019} for other approaches.

\subsubsection{Approximate CPI}

We include a variation of CPI in which the same prediction function,  $f=a(\bZ)$, is used
    compute on both datasets. That is, in this version of the test, a paired t-test is used to compare the samples
    $$(L(f(\tilde{\x}_1),\tilde{y}_1),\ldots,L(f(\tilde{\x}_n),\tilde{y}_n))$$
    and
    $$(L(f(\tilde{\x}^{\pi_1}_1),\tilde{y}_1),\ldots,L(f(\tilde{\x}^{\pi_1}_n),\tilde{y}_n)).$$

This procedure is essentially  the method 
described by \citet{breiman2008} to obtain
p-values for the importance measures produced by random forests, with the exception that   \citet{breiman2008}
use a $z$-test instead.

 \subsubsection{Approximate COINP}
A drawback of COINP is that it is computationally intensive, especially if $a$ is a slow predictive algorithm.
We attempt to overcome these issues by computing the 
rank of 
    $\widehat{R}(a(\bZ),\tilde{\bZ})$ on 
    $$\{\widehat{R}(a(\bZ),\tilde{\bZ}^{\pi_1}),\ldots,\widehat{R}(a(\bZ),\tilde{\bZ}^{\pi_B})\},$$  
    that is, we train $a$ only once (on the original dataset). In other words,
approximate COINP using the same procedure as that described in
Algorithm \ref{alg:perm}, with the exception that line 4 and 5 are now replaced by $f_j \gets f$.
 
\subsection{Simulation study description}
Next,  we describe the details of the simulation study performed  to evaluate 
the proposed method.

\subsubsection{Simulation scenarios}

We generate artificial datasets for our simulation study using various distributions.
We restrict our comparisons to regression settings with the squared loss,
$L(y,\widehat{y})=(y-\widehat{y})^2$, even though our method is general and can be used for classification as well.
Moreover,
in what follows we will
always test conditional independence of a single feature,
i.e., $|S|=1$. 

The first and second scenarios
have features that are independent of each other.

\begin{multicols}{2}
\vfill\null
Distribution 1:
\begin{align*}
& Y_i = \X_i \beta + \epsilon \\
& \beta = (0.7, 0.16, 0.39, \beta_S, 0.75) \\
& \epsilon \sim \text{SKN}(-0.3, 1.1, 2) \\
&  Z_{i,j} \sim \text{SKN}(0, 0.1, 2), \ j=1,\ldots,5 \text{ (i.i.d)} \\
&  X_1 = |Z_1|^{1.3} \\
& X_2 = \cos(Z_2) \\
& X_3 = \log(|Z_1 Z_3|) \\
& X_4 = \log(|Z_3|) \\
&  X_5 = \sqrt{|Z_4|} \\
& \text{Observed input}: (X_1, X_2) \\
& \text{Observed output}: Y
\end{align*}

\columnbreak
\vfill\null
Distribution 2:
\begin{align*}
& Y = Z \beta + \epsilon \\
& \beta \approx (0.7, 0.16, 0.39, \beta_S, 0.75) \\
& \epsilon \sim \text{SKN}(-0.3, 1.1, 2) \\
&  X \sim \text{SKN}(0, 0.1, 2) \text{ (i.i.d)} \\
&  Z_1 = |X_1|^{1.3} \\
& Z_2 = \cos(X_2) \\
& Z_3 = \log(|X_1 X_3|) \\
& Z_4 = \log(|X_3|) \\
&  Z_5 = \sqrt{|X_4|} \\
& \text{Observed input}: \X \\
& \text{Observed output}: Y
\end{align*}
\end{multicols}

In the other settings, we add
correlation to the features
(here SKN stands for skew normal distribution with location, scale and shape parameters respectively):

\begin{multicols}{2}
\vfill\null
Distribution 3:
\begin{align*}
& Y_i = \X_i \beta + \epsilon \\
& \beta = (3, \beta_S) \\
& \X_i \sim N(0, \Sigma) \\
& \epsilon \sim N(0, 0.5) \\
& \Sigma_{0,0} = 1 \\
& \Sigma_{0,1} = 0.9 \\
& \text{Observed input}: \X \\
& \text{Observed output}: Y 
\end{align*}

\columnbreak

Distribution 4:
\begin{align*}
& Y_i = \X_i \beta + \epsilon \\
& \beta = (3, \beta_S) \\
& \epsilon \sim \text{beta}(2, 2) \\
& W_j \sim \text{beta}(1, 1) \text{ for } j \in \{1,2\} \\
& Z \sim N(-0.5, 1) \\
& \X_{i,j} = Z + W_j \text{ for } j \in \{1,2\} \\
& \text{Observed input}: \X \\
& \text{Observed output}: Y 
\end{align*}
\end{multicols}

For all the combinations described above, we  vary $\beta_S$ in $\{0, 0.01, 0.1, 0.6\}$ and the number of observations in $\{1000, 10000\}$. Notice that, in all
settings, $H_0$ holds if, and only if, $\beta_S=0$. Moreover,
as $|\beta_S|$ increases, the 
conditional dependency of $Y$
on $\X^S$ also increases.

For each setting
we run 200 independent tests
to estimate the power of each test. We set $B=100$.

We compare three choices
for the prediction function $a$:
\begin{enumerate}
    \item 
\textbf{Linear regression}, implemented using scikit-learn Python package \citep{scikit-learn}.

\item \textbf{Feedforward neural networks}.
The specification 
of the  network is as follows:
\begin{itemize}
\item \textbf{Optimizer}: we work with the Adamax optimizer \citep{adam-optim} and decrease its learning rate if improvement is seen on the validation loss for a considerable number of epochs.

\item \textbf{Initialization}: we used the initialization method proposed by \citet{nn-initialization}.

\item \textbf{Layer activation}: we chose ELU \citep{elu} as activation function.

\item \textbf{Stop criterion}: a 90\%/10\% split early stopping for small datasets and a higher split factor for larger datasets (increasing the proportion of training instances) and a patience of 50 epochs without improvement on the validation set.

\item \textbf{Normalization and number of hidden layers}: batch normalization, as proposed by \citet{batch-normalization}, is used in this work in order to speed-up the training process, specially since our networks have 5 hidden layers with 100 neurons each.

\item \textbf{Dropout}: here we also make use of dropout which as proposed by \citet{dropout}.

\item \textbf{Software}: we have PyTorch as framework of choice which works with automatic differentiation.
\end{itemize}

\item \textbf{Random Forests}. We use Python's scikit-learn  package with all its default tuning parameters, except for \textit{n\_estimators} (number of trees), which is increased to 300 for better prediction performance.
\end{enumerate}

The Python package and implementation scripts for this work are available at: \url{https://github.com/randommm/nnperm}.

\subsection{Results}

\begin{figure}[!htb]			\begin{center}	\includegraphics[width=0.95\textwidth]{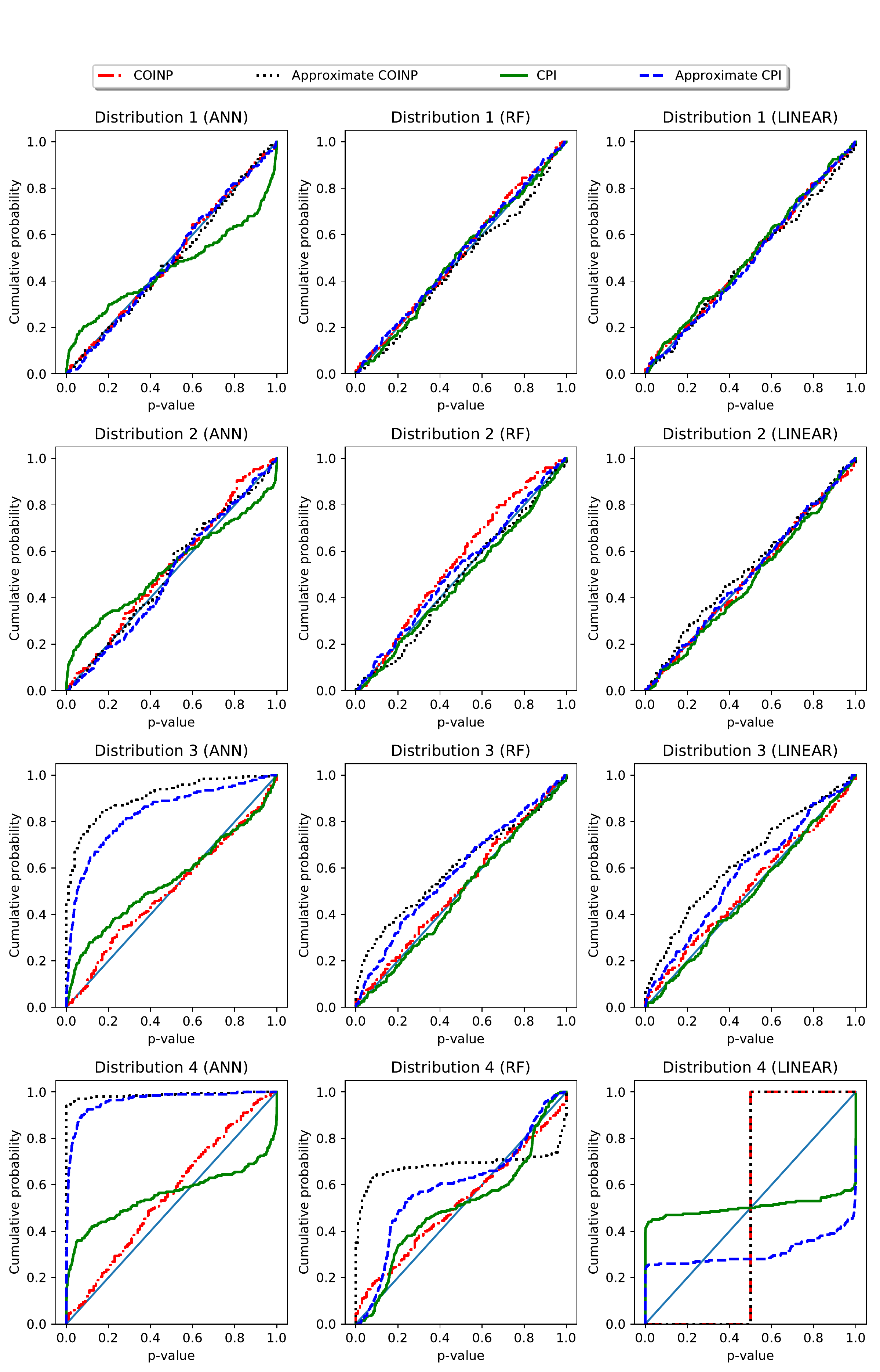}
\caption{Cumulative distribution function of the p-values for the setting with $n=1,000$ under the null hypothesis
(i.e., for choice the $\beta_S=0$).}	\label{fig::uniformity_smalln} \end{center}	\end{figure}

\begin{figure}[!htb]			\begin{center}	\includegraphics[width=0.95\textwidth]{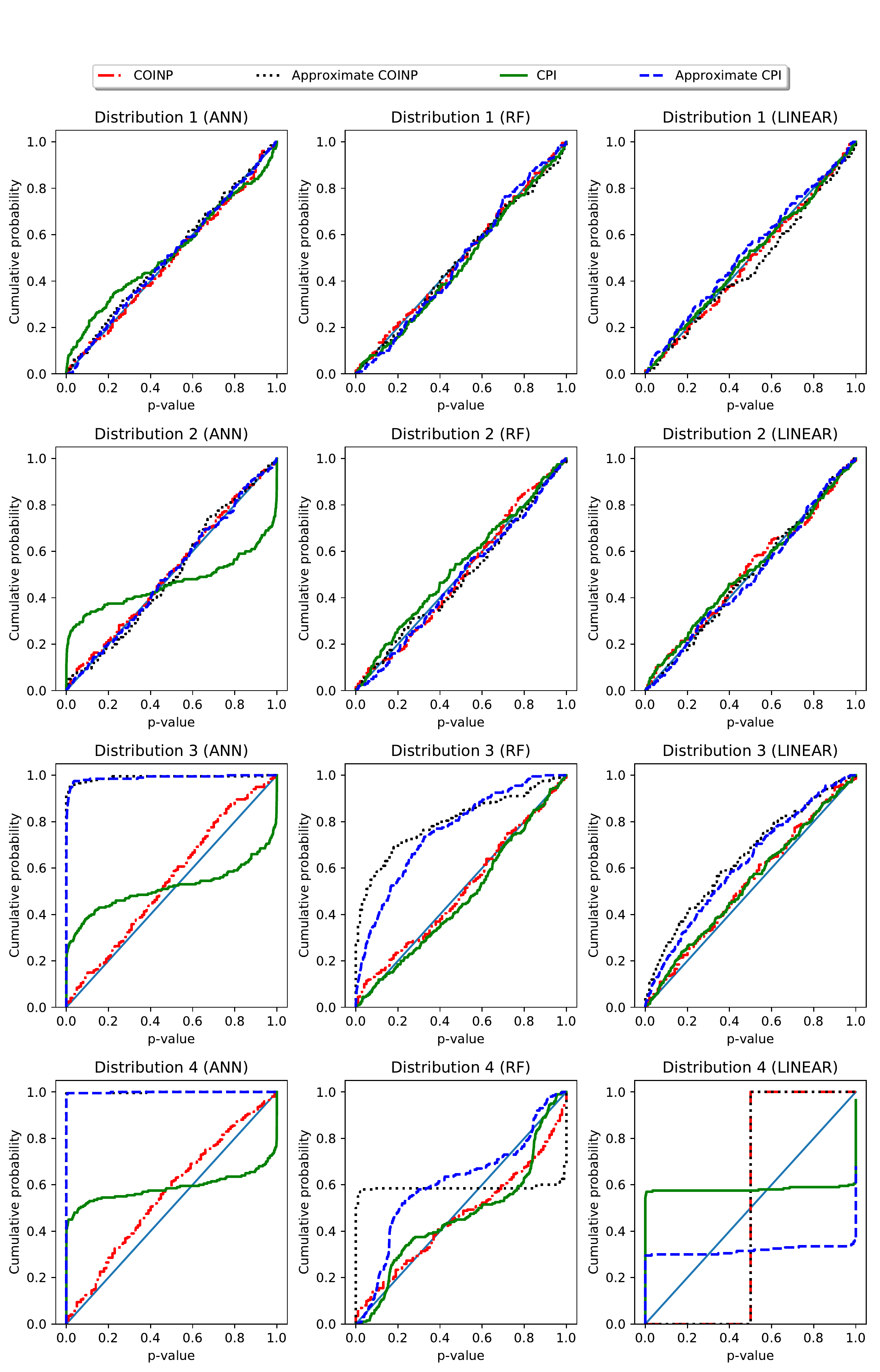}
\caption{Cumulative distribution function of the p-values for the setting with $n=10,000$ under the null hypothesis
(i.e., for choice the $\beta_S=0$).}	\label{fig::uniformity_largen} \end{center}	\end{figure}

Figures \ref{fig::uniformity_smalln}
and \ref{fig::uniformity_largen} show the cumulative distribution functions of the p-values for each of the settings under the null hypothesis (i.e., for  the choice $\beta_S=0$). 
Proper p-values
need to be uniformly distributed under the null hypothesis, and thus the cumulative distribution function should
be close to the 45$^0$ line. 
The figure indicates that approximate methods only lead to proper p-values in settings  1 and 2. These are exactly the cases in which the covariates
are independent of each other 
\izbicki{algum motivo para isso? tinha aquele artigo que falava do breiman testar independencia nao condicional apenas}. Moreover,
the exact methods
come closer to leading to proper p-values in most cases. Exceptions to this are p-values obtained by CPI using artificial neural networks.
This possibly happens because the networks do not converge in some simulations. This leads to extremely large values for the loss functions in some cases, which directly influence the t-test used by CPI.
COINP, on the other hand, is immune to outliers because it relies on the evaluation of \emph{ranks}
as opposed to averages. This in turn guarantees that the distribution of its p-values are closer to uniformity under a larger variety of settings.
We notice that an attempt to get better results for CPI in these settings is to consider the 
logarithm of the loss function, as suggested by
\citet{watson2019}.

\begin{figure}[!htb]			\begin{center}	\includegraphics[width=0.95\textwidth]{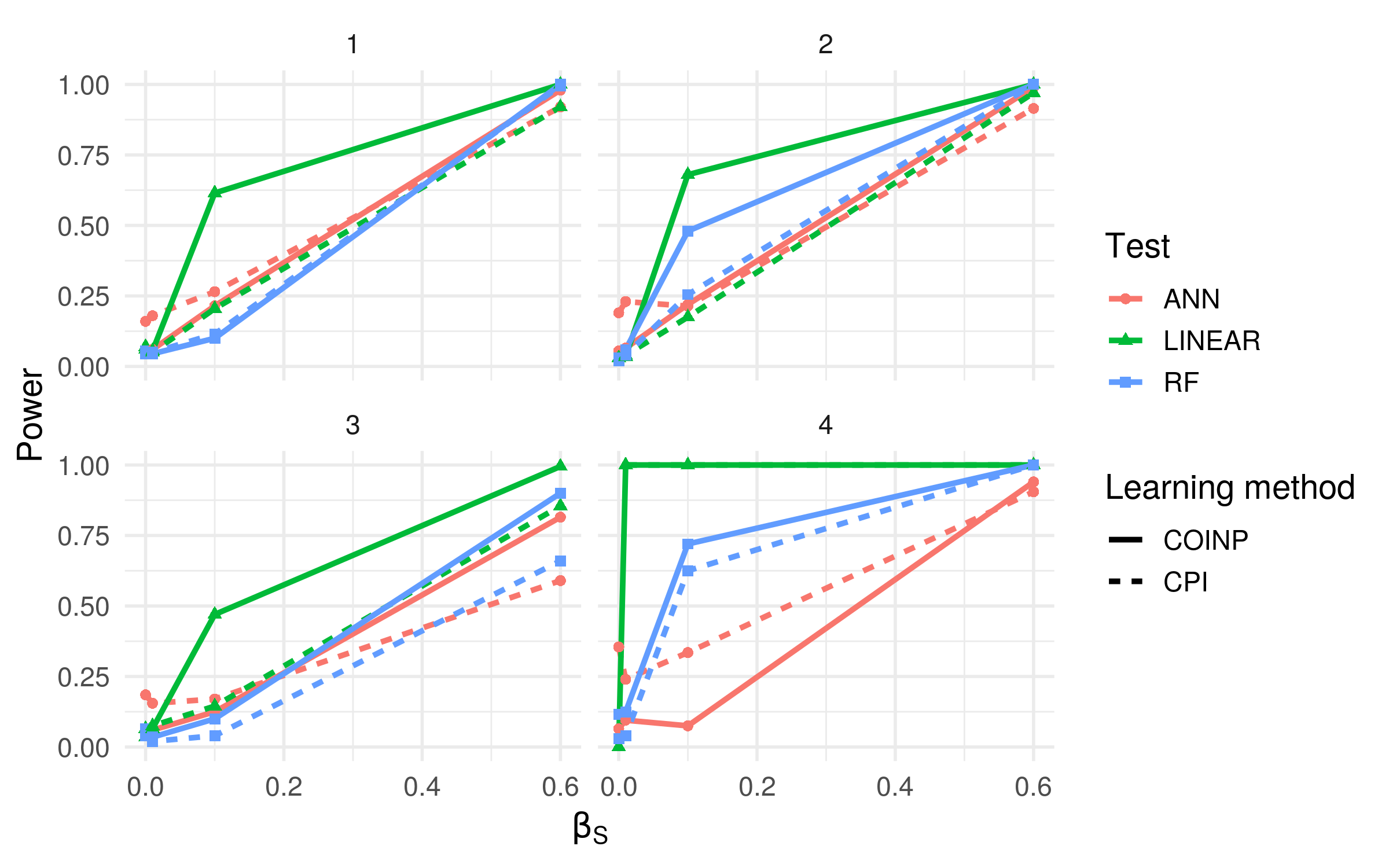}
\caption{Power function for all setting with $n=1,000$ for $\alpha=5\%$.}	\label{fig::power_smalln} \end{center}	\end{figure}

\begin{figure}[!htb]			\begin{center}	\includegraphics[width=0.95\textwidth]{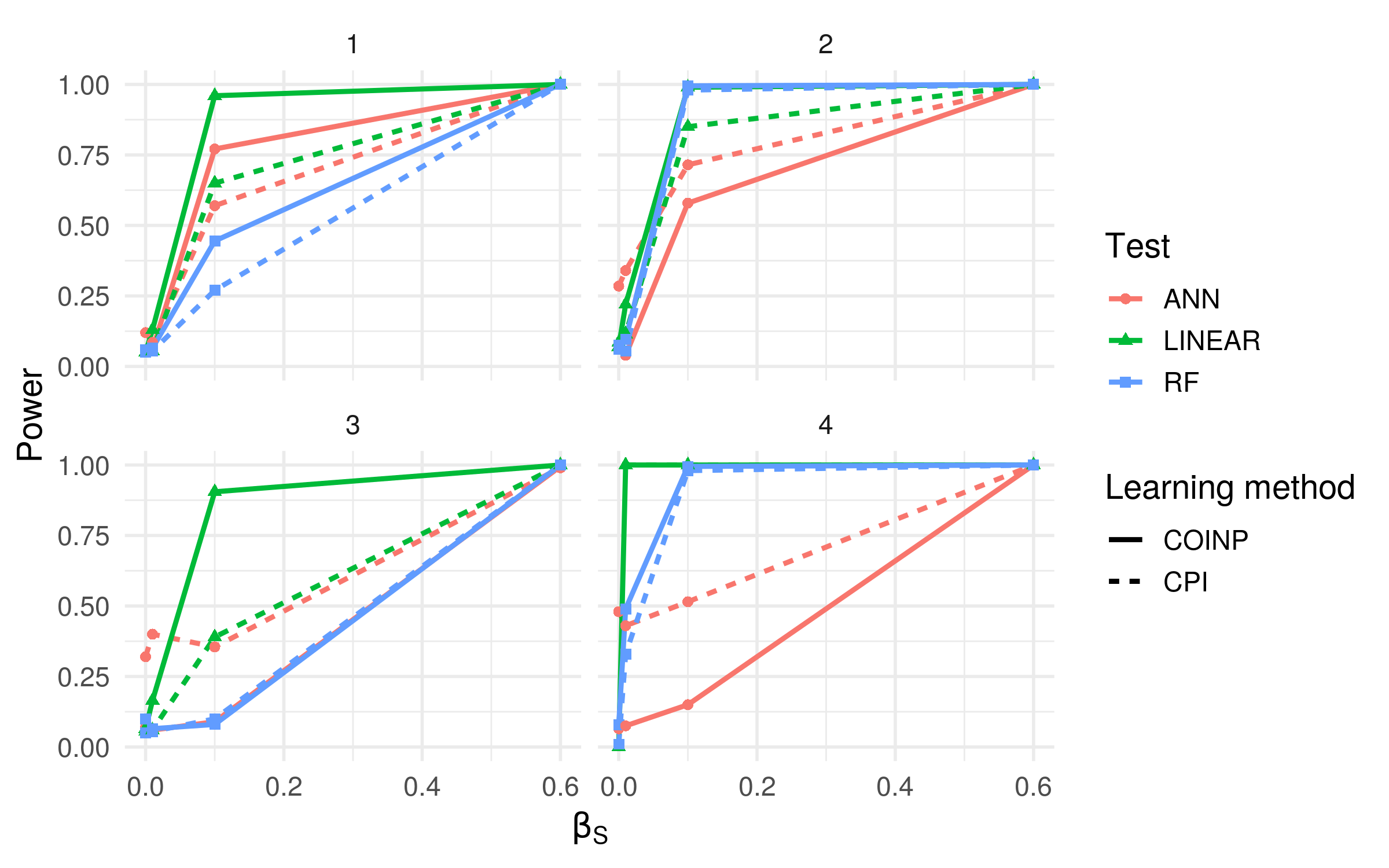}
\caption{Power function for all setting with $n=10,000$ for $\alpha=5\%$.}	\label{fig::power_largen} \end{center}	\end{figure}

Next, we compare the power function of the testing methods. Because only COINP and
CPI had valid p-values, we restrict
the comparisons to these methods.
Figures \ref{fig::power_smalln} and
\ref{fig::power_largen}
show the power of each test as a function of $\beta_S$.
The plot indicates that all procedures achieve higher power as $\beta_S$ increases.
Moreover, in most settings COINP leads to better power than CPI. In these examples,
higher power is achieved when using a linear regression
for COINP.
This can be explain by the fact that in all settings the true nature of the conditional distribution of $Y|\x$ is close to linear.
By comparing the COINP results
from 
Figures \ref{fig::power_smalln} and
\ref{fig::power_largen}, it is also clear the for larger sample sizes (Figure \ref{fig::power_largen}), the power of COINP is larger.
This indicates that the testing procedure is consistent.
\izbicki{o que mais de interessante podemos falar?}

\FloatBarrier
\section{A dataset analysis example}
\label{sec:real_data}
%\izbicki{analisar um problema de predicao usual (calcular pvalor para cada feature talvez usando florestas, junto com a medida de importancia usual)}
As an illustrative example, we take the classical diamonds dataset which is readily available from ggplot2 library and Kaggle. We work with price as the response variable.

In Table \ref{tab:real-data-pvalues}, we present the p-values for each model together with the classical importance measure obtained from random forests.

\begin{table}[!htb]
 \centering
 \caption{P-values for hypothesis testing for each comparison method compared to Random forest traditional importance measures.}
\begin{tabular}{lllllllllll}
\toprule
   &     & carat & depth & table &     x &     y &     z &   cut & color & clarity \\
\midrule

\multirow{4}{*}{ann} & App COINP &  0.00 &  0.00 &  0.00 &  0.00 &  0.00 &  0.00 &  0.00 &  0.00 &    0.00 \\
   & App CPI &  0.00 &  0.02 &  0.03 &  0.00 &  0.00 &  0.00 &  0.00 &  0.00 &    0.00 \\
   & COINP &  0.00 &  0.60 &  0.99 &  0.06 &  0.38 &  0.78 &  0.01 &  0.00 &    0.00 \\
   & CPI &  0.01 &  1.00 &  0.00 &  0.00 &  1.00 &  0.89 &  0.17 &  0.00 &    0.00 \\
\cline{1-11}
\multirow{4}{*}{linear} & App COINP &  0.00 &  0.00 &  0.00 &  0.00 &  0.33 &  0.89 &  0.00 &  0.00 &    0.00 \\
   & App CPI &  0.00 &  0.02 &  0.00 &  0.00 &  0.06 &  0.01 &  0.00 &  0.00 &    0.00 \\
   & COINP &  0.00 &  0.00 &  0.00 &  0.00 &  0.27 &  0.92 &  0.00 &  0.00 &    0.00 \\
   & CPI &  0.00 &  0.02 &  0.07 &  0.00 &  0.00 &  0.01 &  0.03 &  0.00 &    0.00 \\
\cline{1-11}
\multirow{4}{*}{rf} & App COINP &  0.00 &  0.00 &  0.00 &  0.00 &  0.00 &  0.00 &  0.00 &  0.00 &    0.00 \\
   & App CPI &  0.00 &  0.00 &  0.04 &  0.00 &  0.00 &  0.00 &  0.00 &  0.00 &    0.00 \\
   & COINP &  0.00 &  0.05 &  0.00 &  0.04 &  0.00 &  0.02 &  0.00 &  0.00 &    0.00 \\
   & CPI &  0.00 &  0.38 &  0.03 &  0.17 &  0.00 &  0.19 &  0.34 &  0.00 &    0.00 \\

\cline{1-11}
\multicolumn{2}{l}{RF Imp measure} & 0.60 & 0.01 & 0.00 & 0.01 & 0.28 & 0.01 & 0.00 & 0.03 & 0.06 \\
\bottomrule
\end{tabular}
 \label{tab:real-data-pvalues}
\end{table}

%The approximate methods almost never gave a value greater than 5\% for any feature (except for two cases in the linear model), indicating an inability of conditional dependency test to produce meaningful results in practice.

%Other than that, we see that in most cases, the COINP and CPI tests tend to agree and we have generally higher p-values on the ANN when compared to random forests.

%\input{sections/05-theory.tex}

\section{Final remarks} 
\label{sec:final}

We have developed a novel approach for testing conditional independence under a predictive setting. We have shown that the p-values obtained by our approach are proper, and that our hypothesis test has larger power than competing approaches under a variety of settings.

When compared to CPI,
our approach is especially appealing for  small sample sizes, because (i) it does not rely on asymptotic approximations such as those required by the t-test, and (ii) its computational burden is not high in those cases.

\section*{Acknowledgments}

Marco Henrique de Almeida Inacio is grateful for the financial support of CAPES: this study was financed in part by the Coordenação de
Aperfeiçoamento de Pessoal de Nível Superior - Brasil (CAPES) -
Finance Code 001. Rafael Izbicki is grateful for the financial support of FAPESP (2017/03363-8)
and CNPq (306943/2017-4).

\FloatBarrier
\printbibliography

\end{document}